\documentclass[10pt,twocolumn,letterpaper]{article}

\usepackage[pagenumbers]{cvpr} 

\usepackage{graphicx}
\usepackage{amsmath}
\usepackage{amssymb}
\usepackage{booktabs}
\usepackage{multirow}
\usepackage{comment}
\usepackage{makecell}  
\usepackage{pifont}    
\usepackage{float}

\usepackage{capt-of}

\usepackage[T1]{fontenc}
\usepackage[utf8]{inputenc}

\newcommand{\cmark}{\ding{51}}  
\newcommand{\xmark}{\ding{55}}  

\usepackage[hidelinks,breaklinks]{hyperref}

\usepackage[capitalize]{cleveref}
\crefname{section}{Sec.}{Secs.}
\Crefname{section}{Section}{Sections}
\Crefname{table}{Table}{Tables}
\crefname{table}{Tab.}{Tabs.}

\def\cvprPaperID{00000} 
\def\confName{arXiv}
\def\confYear{2026}

\begin{document}

\title{\LARGE VLM-AutoDrive: {\mdseries Post-Training \textbf{V}ision-\textbf{L}anguage \textbf{M}odels for Safety-Critical \textbf{Auto}nomous \textbf{Driv}ing Events}}

\author{
\begin{tabular}{c}
\textbf{Mohammad Qazim Bhat\thanks{Corresponding author} \quad Yufan Huang \quad Niket Agarwal \quad Hao Wang}\\
\textbf{Michael Woods \quad John Kenyon \quad Tsung-Yi Lin \quad Xiaodong Yang}\\
\textbf{Ming-Yu Liu \quad Kevin Xie}\\[2pt]
NVIDIA, California, USA
\end{tabular}
}

\maketitle
\begin{abstract}
The rapid growth of ego-centric dashcam footage presents a major challenge for detecting safety-critical events such as collisions and near-collisions—scenarios that are brief, rare, and difficult for generic vision models to capture. While multimodal large language models (MLLMs) demonstrate strong general reasoning ability, they underperform in driving contexts due to domain and temporal misalignment.

We introduce VLM-AutoDrive, a modular post-training framework for adapting pretrained Vision-Language Models (VLMs) to high-fidelity anomaly detection. The framework integrates metadata-derived captions, LLM-generated descriptions, visual question answering (VQA) pairs, and chain-of-thought (CoT) reasoning supervision to enable domain-aligned and interpretable learning. Off-the-shelf VLMs such as NVIDIA’s Cosmos-Reason1 7B (CR1) exhibit near-zero Collision recall in zero-shot settings; fine-tuning with VLM-AutoDrive improves Collision F1 from 0.00 to 0.69 and overall accuracy from 35.35\% to 77.27\%.

VLM-AutoDrive offers a scalable recipe for adapting general-purpose VLMs to safety-critical, temporally localized perception tasks. Evaluated on real-world Nexar dashcam videos, it achieves substantial gains in Collision and Near-Collision detection while producing interpretable reasoning traces, bridging the gap between perception, causality, and decision reasoning in autonomous driving.

\end{abstract}

\section{Introduction}
\label{sec:intro}

The widespread deployment of dashcams in consumer vehicles has resulted in an enormous volume of ego-centric video data, capturing diverse driving scenarios across real-world environments. These video streams are a rich source of information for downstream applications such as driver behavior analysis, traffic safety monitoring, insurance automation, and autonomous driving system validation. However, detecting critical events such as collisions, near-collisions, and traffic anomalies within these high-frame-rate, untrimmed videos remains a difficult and labor-intensive task.

MLLMs, which integrate visual and textual reasoning into a unified architecture, have shown impressive results across general tasks such as VQA, captioning, and video understanding. Despite these capabilities, pretrained VLMs fail to generalize effectively to the domain of driving anomaly detection. Our experiments show that even advanced models such as CR1~\cite{nvidia2025cosmosreason1physicalcommonsense} demonstrate near-zero precision and recall for the Collision class when evaluated in a zero-shot setting. These models consistently default to predicting “Normal Driving”, highlighting a strong domain gap and a lack of temporal sensitivity.

To address this, we propose a modular and extensible post-training framework to adapt pretrained VLMs for safety-critical, temporally localized driving tasks. Instead of building a model from scratch, our goal is to unlock the capabilities of powerful general-purpose VLMs by exposing them to domain-aligned supervision. The proposed framework comprises the following key components:

\begin{itemize}
    \item \textbf{Empirical analysis of zero-shot limitations} in pretrained VLMs, revealing their ineffectiveness in fine-grained, high-risk driving scenarios.
    \item \textbf{A flexible post-training framework} combining SFT and RL, suitable for domain adaptation using limited labeled data.
    \item \textbf{A scalable supervision pipeline} that generates diverse multimodal signals such as metadata-based captions, synthetic LLM outputs, VQA pairs, and multiple-choice questions (MCQs) to guide model adaptation.
    \item \textbf{Ablation studies identifying key success factors}, such as frame rate sensitivity, data diversity, class balancing, and optimal hyperparameters.
    \item \textbf{A system architecture} that is extensible beyond Collision detection, enabling future expansion to other anomaly classes with minimal retraining effort.
    \item \textbf{Integration into Cosmos Video Curator (CVC):} We aim to make our full annotation pipeline available via CVC, which is already used for video chunking and captioning.
\end{itemize}

Together, our findings underscore the necessity of domain-aligned supervision and post-training to unlock the full potential of VLMs in safety-critical automotive contexts.

Our current system is trained to classify videos into three categories: \textit{Normal Driving}, \textit{Near-Collision}, and \textit{Collision}. However, the architecture is designed to be easily extended to additional anomaly types such as sudden braking, traffic violations, or unsafe lane changes, by updating the supervision pipeline and label set with minimal retraining.
This paper presents our end-to-end framework and evaluates it on a representative dashcam dataset collected from Nexar. We also conduct ablations on data diversity, class imbalance, and frame rate sensitivity to identify key success factors. Our goal is not to propose a novel algorithm, but to provide a practical, reproducible pipeline for adapting VLMs to safety-critical perception tasks in the automotive domain.



\begin{figure}[t]  
    \centering
    \includegraphics[width=\columnwidth]{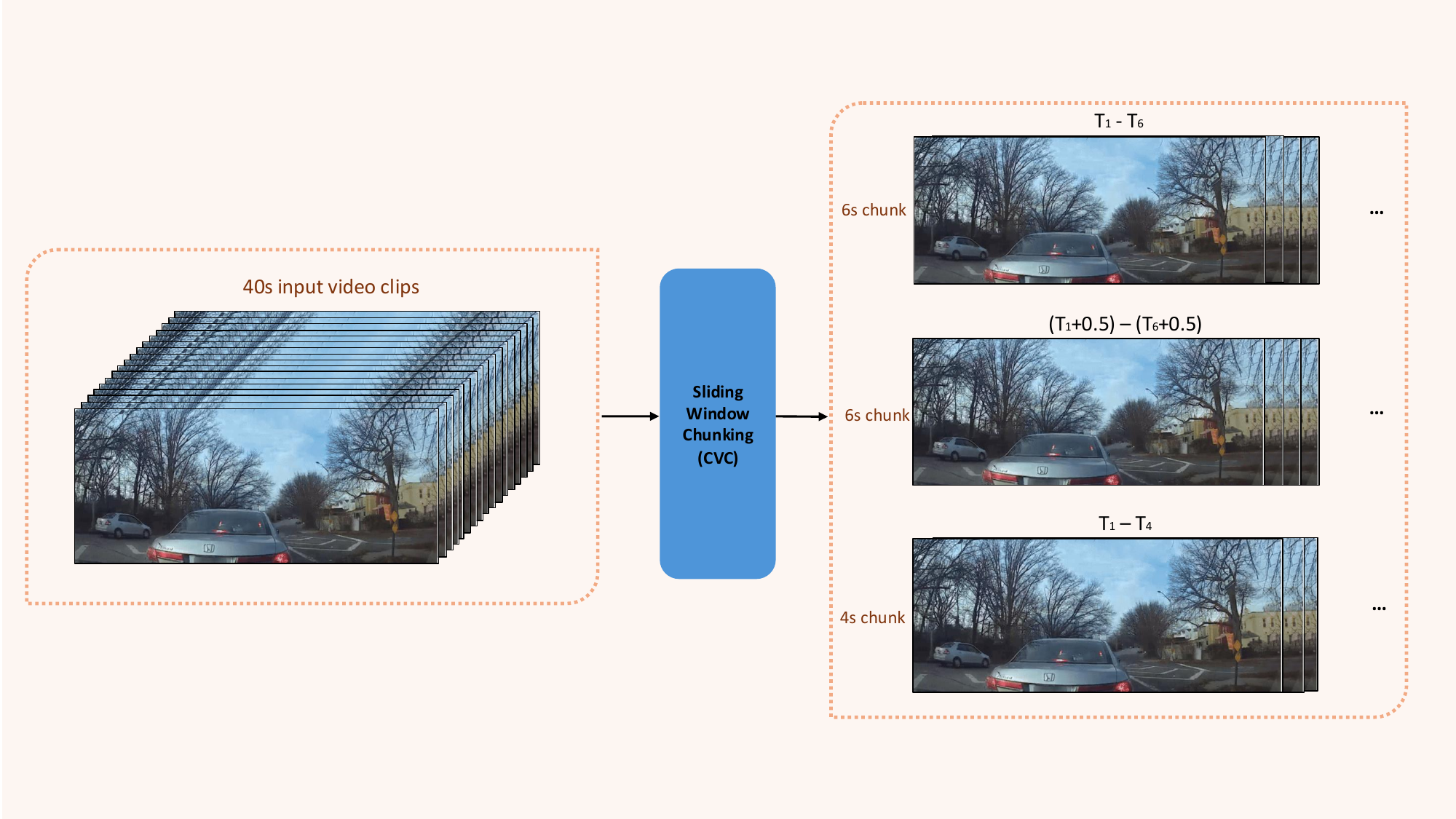}
    \caption{Sliding Window Chunking.}
    \label{fig:column-fig}
\end{figure}





\begin{figure*}[t]
    \centering
    \includegraphics[width=\textwidth]{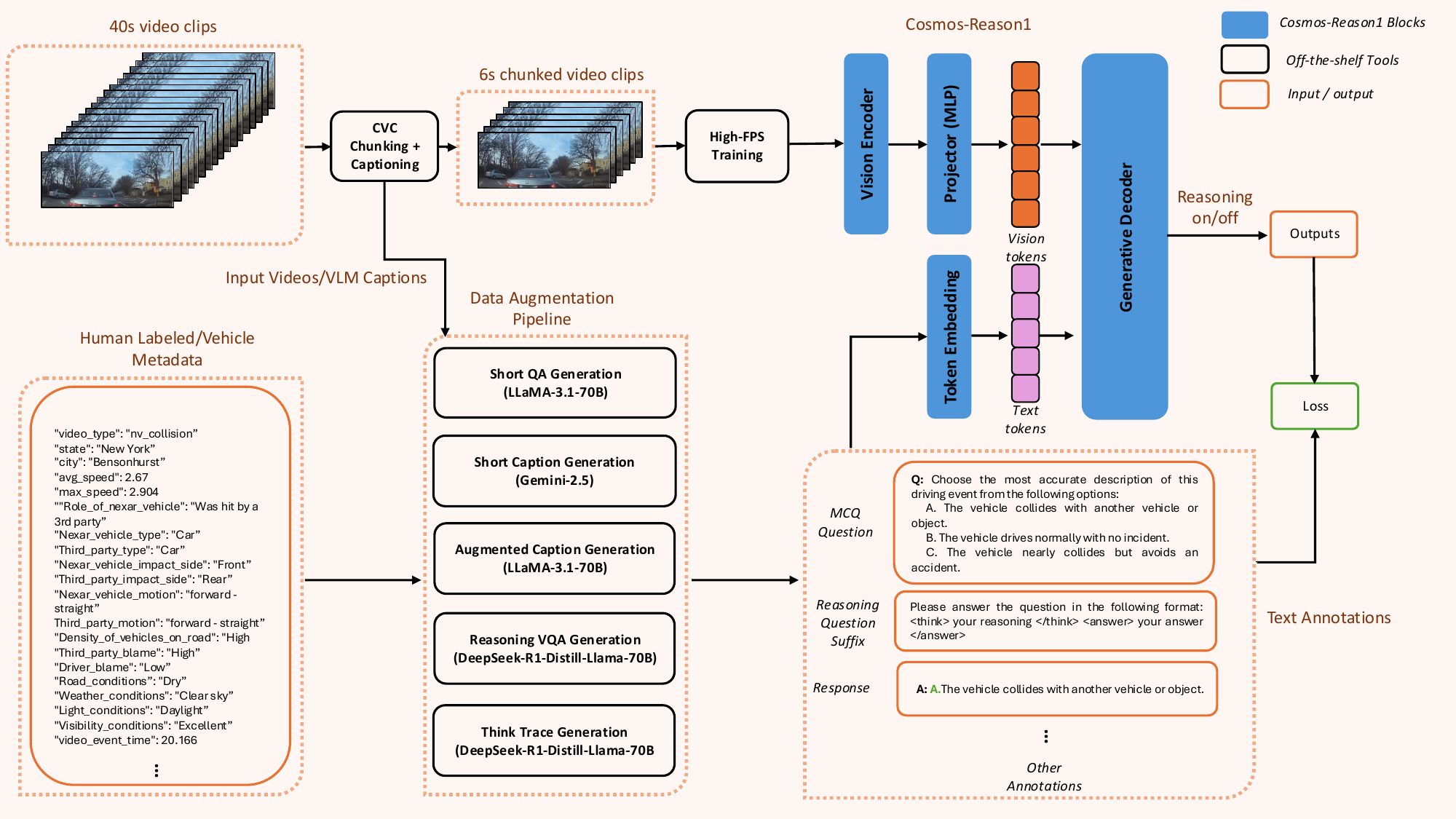}
    \caption{System Diagram.}
    \label{fig:wide}
\end{figure*}

\section{Related Work}

\subsection{VLMs for Video Understanding}

The field of VLMs is seeing a surge in research. The development of closed-source foundation models like GPT-4V~\cite{openai2024gpt4technicalreport} and Gemini~\cite{geminiteam2025geminifamilyhighlycapable} has inspired the research community to develop open-source VLMs and pre-training frameworks, with recent models like LLaVA~\cite{liu2024llavanext}, Qwen-VL~\cite{Qwen-VL}, InternVL~\cite{zhu2025internvl3exploringadvancedtraining}, and NVILA~\cite{liu2025nvilaefficientfrontiervisual} leading the way. These models are being applied in various scenarios, from robotics to autonomous driving. 

Inspired by reasoning LLMs like DeepSeek-R1~\cite{deepseekai2025deepseekr1incentivizingreasoningcapability}, recent work has focused on developing reasoning VLMs. Models such as LLaVA-CoT~\cite{xu2024llavacot}, CR1~\cite{nvidia2025cosmosreason1physicalcommonsense} and Long-RL~\cite{longrl} aim to build a chain-of-thought to enable more complex reasoning. NVIDIA's CR1~\cite{nvidia2025cosmosreason1physicalcommonsense} is a prominent example of a VLM post-trained on data requiring physical common sense and embodied reasoning, enabling it to output a chain-of-thought reasoning before providing an answer.

\subsection{VLMs in Autonomous Driving }

VLMs are increasingly being integrated into autonomous driving systems to enhance reasoning, interpretability, and generalization. By unifying perception and language understanding in a shared embedding space, VLMs demonstrate improved zero-shot performance across diverse driving conditions and rare scenarios. 

Prior work on safety-critical driving-video understanding includes dashcam anomaly benchmarks (DoTA~\cite{haresh2020anomalydetectiondashcamvideos}) and risk-aware captioning/localization (DRAMA~\cite{malla2022dramajointrisklocalization}).

Early approaches like GPT-Driver~\cite{mao2023gptdriverlearningdrivegpt} show that frozen VLMs can generate control plans and rationales from visual and textual inputs, but they suffer from poor spatial grounding and ambiguous outputs. To overcome these issues, recent works such as BEVDriver ~\cite{winter2025bevdriverleveragingbevmaps}, TOKEN ~\cite{jiang2025tokenefficientlongvideounderstanding}, and Sce2DriveX ~\cite{zhao2025sce2drivexgeneralizedmllmframework} enhance spatial understanding. DriveVLM ~\cite{tian2024drivevlmconvergenceautonomousdriving} framework employs a chain‑of‑thought process including scene description, analysis, and hierarchical planning and introduces a hybrid DriveVLM‑Dual ~\cite{tian2024drivevlmconvergenceautonomousdriving} system that combines VLM reasoning with traditional 3D perception for improved spatial grounding and real‑time control in AV scenarios. More recently, CR1 ~\cite{nvidia2025cosmosreason1physicalcommonsense} extends VLM capabilities to the autonomous driving domain by incorporating physical common sense and embodied reasoning through chain-of-thought supervision. It enables models to produce interpretable, multi-step rationales for complex driving scenarios, improving temporal and causal understanding.

Based on Cosmos-Reason, Alpamayo-R1~\cite{nvidia2026alpamayor1bridgingreasoningaction} tightly integrates explicit reasoning with trajectory planning to better handle rare, safety-critical scenarios. It introduces a causally grounded reasoning dataset and a modular vision-language-action system trained with both supervised learning and reinforcement learning to align reasoning with actions.

\section{Methodology}

We propose a modular, post-training framework to adapt pretrained VLMs for detecting anomalous driving events in ego-centric dashcam footage. The design is guided by two observations: (1) general-purpose VLMs underperform in this domain due to a lack of task-specific supervision and sensitivity to short-duration events, and (2) adapting powerful pretrained models is more efficient and scalable than training from scratch. Our framework emphasizes data-centric adaptation, using diverse, domain-aligned supervision signals to bridge the gap between open-domain reasoning and high-fidelity automotive perception.

\subsection{Problem Setup}

We formulate the task as short-duration video classification for safety-critical driving events. Given an ego-centric dashcam video clip of 4--6 seconds, the objective is to assign one of three driving event labels:

\begin{itemize}
    \item \textbf{Normal Driving}: No incident occurs.
    \item \textbf{Near-Collision}: A close call where the ego vehicle narrowly avoids impact.
    \item \textbf{Collision}: A physical contact involving the ego vehicle and another vehicle, pedestrian, or object.
\end{itemize}

These clips are extracted from longer dashcam videos using a sliding-window chunking strategy. Critical events often span just 5--15 frames, requiring high frame rates and fine-grained temporal sensitivity for reliable detection.

Despite the simplicity of the classification task, several challenges make it uniquely difficult in real-world settings.

\subsubsection{Key Challenges}

\begin{itemize}
    \item \textbf{High Temporal Fidelity}: The Collision and Near-Collision classes are extremely brief, often visible in fewer than 0.5 seconds of footage, necessitating fine-grained temporal modeling at 30 FPS or higher.
    
    \item \textbf{Severe Class Imbalance}: Real-world dashcam data is dominated by uneventful driving, making anomalous events such as collisions extremely rare. This bias severely hampers model generalization to safety-critical cases.

    \item \textbf{Sparse Multimodal Supervision}: Raw labels alone are insufficient for adapting general-purpose VLMs. Rich, task-specific supervision such as captions, VQA, and reasoning traces are essential for learning causal and contextual cues.
\end{itemize}

\subsection{VLM-AutoDrive System Components}

We propose \textbf{VLM-AutoDrive}, a modular post-training framework for adapting general-purpose VLMs to safety-critical driving anomaly detection. Rather than training from scratch, we expose pretrained VLMs to a rich, task-aligned supervision pipeline that enhances their temporal sensitivity, causal reasoning, and safety-critical classification ability.
Figure~\ref{fig:wide} provides an overview of the full system pipeline.

\paragraph{Overview.}
The VLM-AutoDrive pipeline consists of three main components:

\begin{itemize}
    \item \textbf{Input Source}: 40-second ego-centric dashcam videos, manually labeled with event annotations and augmented with structured metadata (e.g., vehicle motion, impact side, blame).
    
    \item \textbf{Multimodal Supervision Generation}: We convert raw inputs into diverse supervision signals including MCQs, captions, VQA pairs, and reasoning traces, to guide supervised fine-tuning (SFT). These annotations are synthesized using a combination of rule-based templates, LLM prompting (e.g., LLaMA-3.1, Gemini), and video captioning VLMs (e.g., NVILA, CR1).

    \item \textbf{Fine-tuning Strategy}: The synthesized dataset is used to post-train models via SFT, optionally followed by reinforcement learning (RL) to further align reasoning behavior with driving-specific objectives.
\end{itemize}

\begin{figure*}[t]
    \centering
    \includegraphics[width=\textwidth]{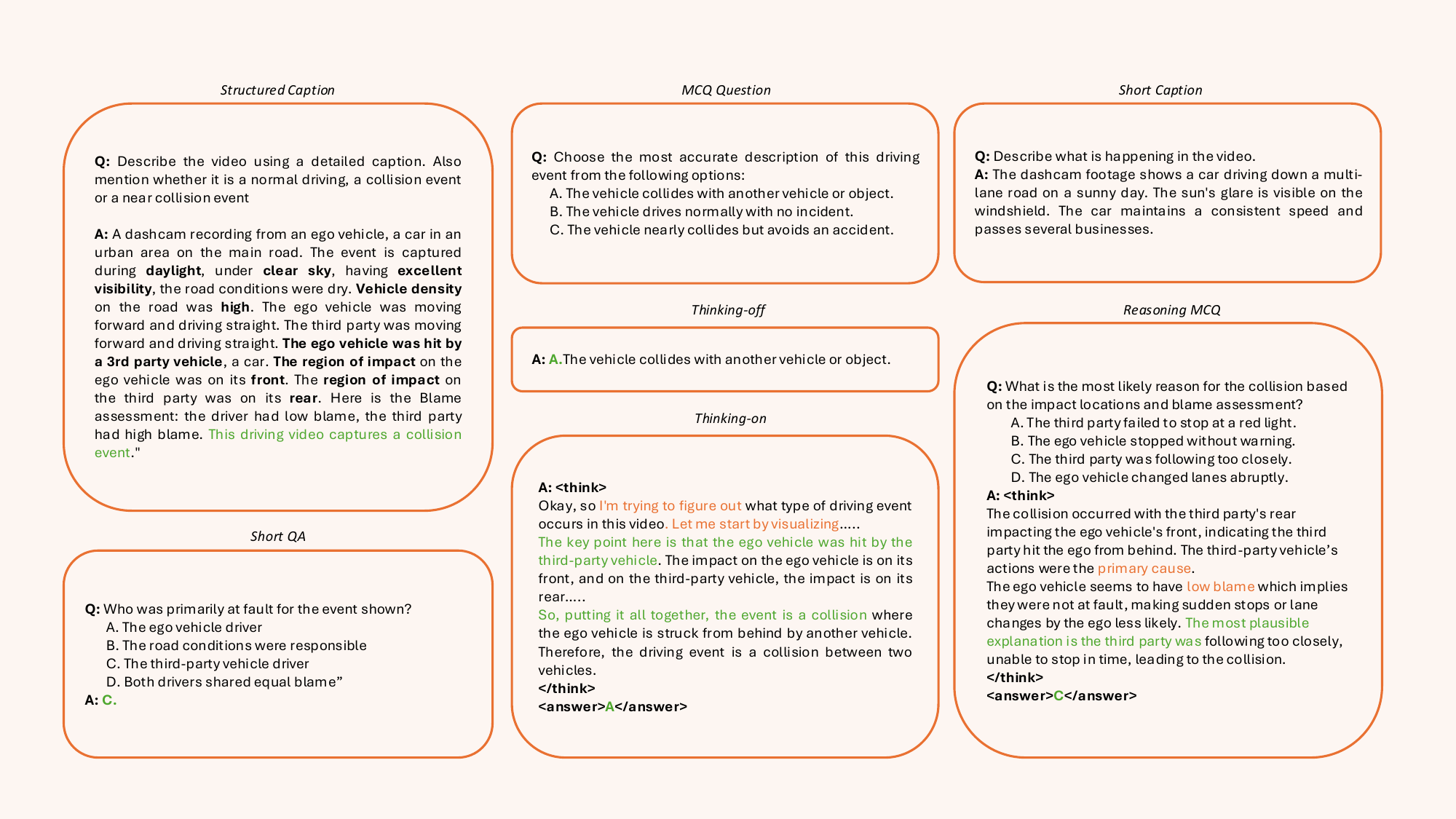}
    \caption{Data Pipeline Examples.}
    \label{fig:data_annotation_example}
\end{figure*}

We begin by framing driving event classification as a supervised MCQ task.  
From the chunked dataset, we construct approximately 53{,}000 MCQ samples, split into 45{,}000 training samples. From the remaining samples, we use 198 samples (66 each for the three classes) as our test set. 
Each MCQ presents a short video clip (or its textual description) and asks the model to choose the correct event label from three options.  
This serves as the foundation for subsequent multimodal augmentations (see Figure~\ref{fig:data_annotation_example}).

Building on this baseline, we generate richer annotations through a multi-stage process:

\begin{itemize}
    \item \textbf{Caption Generation:} We use VLMs such as Gemini~\cite{geminiteam2025geminifamilyhighlycapable} and NVILA~\cite{liu2025nvilaefficientfrontiervisual} to generate generic visual captions. 
    
    \item \textbf{Metadata-to-Text Conversion:} Structured metadata which includes scene context, incident details, and blame assessment is programmatically converted into detailed textual descriptions.
    
    \item \textbf{VQA Pair Generation:} We use LLaMA-3.1-70B~\cite{grattafiori2024llama3herdmodels} to generate spatial, environmental, and causal questions and answers based on the metadata-derived captions.
    
    \item \textbf{Reasoning MCQ Creation:} Finally, we use DeepSeek-R1-Distill-LLaMA-70B~\cite{deepseekai2025deepseekr1incentivizingreasoningcapability} to create chain-of-thought MCQs, preserving and enhancing causal reasoning capabilities in the fine-tuned CR1~\cite{nvidia2025cosmosreason1physicalcommonsense}.
\end{itemize}

Further details of the prompting pipeline are provided in Section~\ref{sec:dataset}, with full templates in the Appendix.

\subsubsection{Video Chunking and Captioning}

\label{sec:chunking}
The raw dataset consists of approximately 10{,}000 40-second ego-centric videos from the Nexar dashcam platform~\cite{moura2025nexardashcamcollisionprediction}, including around 1{,}000 labeled as Collision and 9{,}000 as Near-Collision clips.

To address the short temporal span of anomalies and the limited context window of VLMs, we apply \textbf{CVC} to segment each video into shorter 4--6 second clips. This \emph{sliding-window chunking} strategy serves two key purposes:

\begin{itemize}
    \item It increases the effective frame rate presented to the model during training, enabling better detection of fine-grained, temporally localized events.
    
    \item It expands the dataset to approximately 53{,}000 annotated clips, with each original video yielding 6--7 overlapping segments and only one of which typically contains the anomalous event. The rest are labeled as normal driving.
\end{itemize}

In addition to chunking, CVC performs short caption generation for each video segment using off-the-shelf video captioning VLMs~\cite{nvidia2025cosmosworldfoundationmodel}, providing initial visual summaries for further annotation steps.

\subsection{Dataset and annotations}

\label{sec:dataset}

Our dataset is sourced from real-world dashcam footage collected by the Nexar platform. It consists of approximately 10{,}000 ego-centric videos, each 40 seconds long, annotated with high-level driving event labels and fine-grained structured metadata (e.g., vehicle roles, motion, impact sides, blame assessments, and environmental conditions).

To enable fine-grained anomaly detection, we apply the chunking strategy described in Section~\ref{sec:chunking}, which expands the dataset to $\sim$53{,}000 short clips of 4--6 seconds each. These chunks serve as the primary units for all annotation and model input.

The resulting dataset exhibits a significant class imbalance with 43{,}000 \textbf{Normal Driving}, 9{,}000 \textbf{Near-Collision}, and 1{,}000 \textbf{Collision} examples.

This skew reflects real-world driving but presents challenges for learning safety-critical behaviors. In Section~\ref{sec:data-mixing}, we describe our balancing strategy to mitigate this issue.

Each chunk is further enriched through a multi-stage annotation pipeline (Figure~\ref{fig:data_annotation_example}), which generates MCQs, structured and LLM-generated captions, VQA pairs, and chain-of-thought reasoning traces. These annotations form the foundation of our supervised fine-tuning dataset described in Table~\ref{tab:data_mixture}.

\subsubsection{Human-Labeled Annotations and Metadata Integration}

Each 40-second video includes human-annotated event labels and fine-grained metadata, capturing attributes such as vehicle roles, motion patterns, impact sides, blame assignment, and environmental context. This metadata provides structured semantic cues beyond simple class labels, forming the basis for generating diverse textual annotations used in downstream supervision.

\subsubsection{Multiple-Choice Question (MCQ) Data}

We frame event classification as a supervised multiple-choice task. From the chunked dataset, we construct $\sim$53{,}000 MCQ samples. This format serves as the foundation for subsequent caption, VQA, and reasoning augmentations (Figure~\ref{fig:data_annotation_example}).

\subsubsection{Caption Augmentation}

To provide richer visual context, we generate two types of captions:

\begin{itemize}
    \item \textbf{Short Captions:} Generic descriptions generated by Gemini-2.5~\cite{geminiteam2025geminifamilyhighlycapable}, applied to each chunked clip, yielding $\sim$50{,}000 samples.

    \item \textbf{Metadata-Derived Captions:} Structured, scene-specific descriptions programmatically created from Nexar metadata, encoding key details like motion, impact side, blame, and weather. This produces $\sim$20{,}000 high-precision captions.
\end{itemize}

\subsubsection{Visual Question Answering (VQA) Augmentation}

From the structured captions, we use LLaMA-3.1-70B~\cite{grattafiori2024llama3herdmodels} to generate $\sim$140{,}000 VQA pairs that include approximately seven targeted questions per clip. These questions span spatial, environmental, causal, and blame-related aspects (Appendix Fig.~\ref{fig:understanding_VQA}). VQA supervision complements the MCQ format by encouraging fine-grained visual reasoning rather than classification alone.

\subsubsection{Reasoning and Think Trace Generation}

Post-training without explicit reasoning supervision often
leads to catastrophic forgetting of pretrained chain-of-thought capabilities. To prevent this, we incorporate reasoning traces using DeepSeek-R1-Distill-LLaMA-70B~\cite{deepseekai2025deepseekr1incentivizingreasoningcapability}. These traces explicitly encode step-by-step reasoning, preserving causal reasoning performance.

\vspace{0.5em}
\noindent We generate two types of reasoning data:

\begin{itemize}
    \item \textbf{Reasoning MCQs:} We augment MCQ samples with think traces, model-generated justifications preceding the final answer. VLM-generated caption, enriched detailed caption, and a classification question for each sample is fed into a reasoning LLM, DeepSeek-R1 in this case. Only the samples with correct output answers from the LLM are retained, resulting in $\sim$12{,}000 reasoning-augmented MCQs (Appendix Fig.~\ref{fig:chunking-full-1}).

    \item \textbf{General Reasoning VQA:} To broaden reasoning beyond classification, we generate open-ended question-answer-reason triplets~\cite{longrl}. DeepSeek-R1 is prompted with a combination of VLM and metadata-derived captions to generate explanations about context, causality, and intent (Appendix Fig.~\ref{fig:think_trace_general_VQA}).
\end{itemize}

\begin{table*}[t]
    \centering
    \caption{Summary of Data Mixture for SFT grouped into Understanding and Reasoning splits}
    \begin{tabular}{llcc}
        \toprule
        \textbf{Category} & \textbf{Annotation Type} & \textbf{Approx. Count} & \textbf{Source/Description} \\
        \midrule
        \multirow{5}{*}{\textbf{Understanding}} 
            & MCQ Data (Unbalanced) & $\sim$53,000 & From Nexar videos - 1k Collision, 9k Near-Collision, 43k Normal \\
            & MCQ Data (Upsampled)  & $\sim$68,000 & Balanced (11k Collision, 14k Near-Collision, 43k Normal) \\
            & Detailed Caption Data & $\sim$20,000 & Rule-based, using Nexar video metadata \\
            & Gemini Caption Data   & $\sim$50,000 & From chunked Nexar videos using Gemini \\
            & VQA Data              & $\sim$124,000 & From detailed captions via LLaMA-3.1-70B \\
        \midrule
        \multirow{3}{*}{\textbf{Reasoning}} 
            & AV Reasoning Data     & $\sim$12,000 & From CR1 SFT stage \\
            & MCQ Reasoning Data     & $\sim$12,000 & From Detailed and VLM captions using DeepSeek R1\\
            & VQA Reasoning Data     & $\sim$10,000 & From Detailed and VLM captions using DeepSeek R1 \\
            
        \midrule
        \multicolumn{2}{l}{\textbf{Total Approx. Count}} & $\sim$349,000 & -- \\
        \bottomrule
    \end{tabular}
    \label{tab:data_mixture}
\end{table*}

\subsubsection{CR1 SFT AV Reasoning Data}

In addition to the MCQ, caption, and reasoning datasets, we incorporated approximately 12,000 autonomous-vehicle (AV) reasoning samples from the SFT stage of the original CR1 training pipeline. These samples contain chain-of-thought reasoning traces tailored for embodied reasoning in driving scenarios.

\subsubsection{Final Dataset Mixture}
The initial MCQ dataset exhibited a pronounced class imbalance, with $\sim$43{,}000 Normal Driving, $\sim$9{,}000 Near-Collision, and $\sim$1{,}000 Collision cases. To mitigate this, Collision and Near-Collision samples were up-sampled by factors of 15$\times$ and 2$\times$, respectively, yielding a balanced set of $\sim$11{,}000 Collision, $\sim$14{,}000 Near-Collision, and $\sim$43{,}000 Normal Driving cases for primary training.

For optimal performance, we curated a multi-source mixture of approximately 349{,}000 annotations, integrating diverse supervision signals across modalities such as MCQ, detailed captions, VLM-generated captions (Gemini), VQA, and reasoning data. The complete breakdown, including counts, origins, and annotation methods, is summarized in Table~\ref{tab:data_mixture}.

\subsubsection{Post-Training Strategy}

Our initial evaluation revealed that zero-shot models failed to detect high-temporal-fidelity events, achieving 0\% recall on Collision cases and overall accuracy close to random chance. Most VLMs, including CR1~\cite{nvidia2025cosmosreason1physicalcommonsense}, even with reasoning enabled, consistently misclassified all events as Normal Driving. This pronounced domain gap highlighted the need for targeted post-training to adapt models to ego-centric driving incident detection.

\textbf{Supervised Fine-Tuning (SFT):} We adapt the base VLM by fine-tuning it with curated multimodal supervision signals, formatted as instruction-style prompts for classification and VQA objectives. Video–text pairs are tokenized, paired with corresponding labels, and optimized to align the model’s perception and reasoning capabilities with the driving anomaly detection task.

\subsection{Model Selection}

Our primary model in focus is \texttt{CR1} ~\cite{nvidia2025cosmosreason1physicalcommonsense} because it is explicitly post-trained for embodied reasoning and physical common sense, unlike most general-purpose VLMs. We also used \texttt{NVILA-8B Video} ~\cite{liu2025nvilaefficientfrontiervisual} and \texttt{Qwen2.5-VL-7B} ~\cite{Qwen-VL}.

\subsubsection{CR1 Foundation Model}

CR1 is a multimodal large language model developed by NVIDIA for ``Physical AI'' tasks. It features a 7-billion-parameter transformer backbone capable of processing both visual and textual inputs. A vision transformer encoder extracts features from images or video frames, which are then projected into the text embedding space before being passed to the LLM decoder. The decoder processes these aligned visual and textual tokens to produce a response.

A key strength of CR1 is its post-training on datasets requiring physical commonsense and embodied reasoning, enabling it to generate chain-of-thought explanations before delivering an answer. This makes it particularly well-suited for interpreting driving scenes, where understanding physics and temporal dynamics is critical. 

\subsection{Extensibility}

The proposed framework is modular, enabling the addition of new driving anomaly types with minimal retraining. Introducing a new class (e.g., \textit{Sudden Stop}) involves:
\begin{itemize}
    \item Updating the metadata parser and caption templates
    \item Generating targeted VQA/MCQ samples via LLM prompts
    \item Expanding the label set and post-training on the augmented dataset
\end{itemize}

This design ensures scalability as new events and datasets emerge. The framework can capture a wide range of anomalies such as red light violations, stop sign violations, illegal turns, no-entry violations, sharp cornering, hard acceleration, and hard braking, by making minor adjustments to the data pipeline. Several of these classes (e.g., sharp cornering, hard acceleration/braking) already have human-annotated ground truth and will be integrated into future post-training data recipes.

\subsection{Implementation Details}

We use NVIDIA CR1 as the base VLM. Post-training is performed on a mixture of human-labeled and LLM-augmented annotations derived from Nexar videos. Performance is evaluated using precision, recall, and F1 score across the three primary classes, with ablation studies measuring the impact of each supervision signal.

\subsection{Key Factors for Success}

Ablation studies identified the following as critical to performance improvements:

\begin{itemize}
    \item \textbf{Diverse supervision:} Combining MCQ, captions, and open-ended QA enhanced generalization.
    \item \textbf{Balanced class distribution:} Prevented bias toward the dominant ``Normal Driving'' class.
    \item \textbf{High temporal fidelity:} 30 FPS and high-resolution video were essential for event capture.
    \item \textbf{Optimized learning rate:} $1 \times 10^{-5}$ yielded stable and effective training.
    \item \textbf{SFT over zero-shot:} SFT significantly improved Collision detection.
    \item \textbf{Reasoning-enriched data:} Incorporating think-trace reasoning data preserved and enhanced causal reasoning capabilities, preventing catastrophic forgetting during SFT.

\end{itemize}

\section{Experiments and Results}

Our experiments involved extensive fine-tuning of \texttt{CR1}, \texttt{NVILA-8B Video}, and \texttt{Qwen2.5-VL-7B}.

\subsection{Experimental Setup}

\begin{table*}[t]
  \centering
  \caption{Per-class precision, recall, and F1-score for different models and settings. Model-A includes both zero-shot and fine-tuned performance.}
  \small
  \begin{tabular}{l|ccc|ccc|ccc|c}
    \toprule
    \textbf{Model} 
    & \multicolumn{3}{c|}{\textbf{Collision}} 
    & \multicolumn{3}{c|}{\textbf{Near-Collision}} 
    & \multicolumn{3}{c|}{\textbf{Normal Driving}} 
    & \textbf{Accuracy} $\uparrow$ \\
    
    \cmidrule(lr){2-4} \cmidrule(lr){5-7} \cmidrule(lr){8-10}
    & \textbf{Prec.} $\uparrow$ & \textbf{Rec.} $\uparrow$ & \textbf{F1} $\uparrow$
    & \textbf{Prec.} $\uparrow$ & \textbf{Rec.} $\uparrow$ & \textbf{F1} $\uparrow$
    & \textbf{Prec.} $\uparrow$ & \textbf{Rec.} $\uparrow$ & \textbf{F1} $\uparrow$
    & \\
    
    \midrule
    Cosmos-Reason1 Zero-shot   & 0.000  & 0.000  & 0.000   & 0.454  & 0.0075  & 0.129   & 0.347 & \textbf{0.984} & 0.513  & 35.35 \\
    Cosmos-Reason1 Fine-tuned  & \textbf{0.947} & 0.545 & 0.692  & 0.696 & 0.833 & 0.758 & 0.765 & 0.939  & 0.843  & 77.27 \\
    \midrule
    NVILA Zero-shot   & 0.000  & 0.000  & 0.000   & 0.250  & 0.500  & 0.333   & 0.500 & 0.667 & 0.571  & 38.89 \\
    NVILA Fine-tuned                       & 0.901 & \textbf{0.696} & \textbf{0.786}  & \textbf{0.782} & \textbf{0.924} & \textbf{0.847}  & \textbf{0.927} & 0.969  & \textbf{0.948}  & \textbf{86.36} \\
    \bottomrule
  \end{tabular}
  \label{tab:final_model_comparison}
\end{table*}
\subsubsection{Dataset}

We evaluate our framework on the $\sim$53{,}000 chunked clips derived from the Nexar dashcam dataset using our annotation pipeline (Section~\ref{sec:dataset}) on classification score on the three categories. These clips are 4--6 seconds long and include metadata-derived captions, visual descriptions, VQA pairs, and reasoning-based MCQs.

To address class imbalance, we upsampled Collision and Near-Collision events by 15$\times$ and 2$\times$, respectively. For training, we use a curated subset that mixes multiple annotation types per clip. The final distribution is summarized in Table~\ref{tab:data_mixture}.

\subsubsection{Training Details}

Model training is performed via SFT, including CoT-SFT. To support high-temporal-fidelity anomaly detection, each video is processed as 180 frames at 30 FPS. The input resolution is adjusted automatically according to the available GPU memory and token budget. We use the AdamW optimizer with a learning rate of $1 \times 10^{-5}$, a batch size of 8, and weight decay of 0.01. Models are trained for 1 epoch.

To handle large model sizes and extended video-text input lengths, we enable BF16 mixed precision, gradient checkpointing, and DeepSpeed ZeRO-3 optimization. All experiments are conducted on a cluster of 32 NVIDIA H100 GPUs.

\subsection{Metrics}

We evaluated performance using per-class \textbf{Precision}, \textbf{Recall} and \textbf{F1-score}, with an emphasis on improving these metrics for the minority classes (Collision and Near-Collision). We also assessed overall accuracy and binary anomaly detection accuracy (normal vs. anomalous driving). For our evaluation benchmark, we randomly sample 66 examples from each of the 3 classes from the test set.

Precision measures the proportion of correctly predicted positive instances out of all instances predicted as positive, indicating the model's ability to avoid false positives:
\[
\text{Precision} = \frac{TP}{TP + FP}
\]
where \(TP\) is the number of true positives and \(FP\) is the number of false positives.

Recall measures the proportion of correctly predicted positive instances out of all actual positive instances, indicating the model's ability to detect all relevant cases:
\[
\text{Recall} = \frac{TP}{TP + FN}
\]
where \(FN\) is the number of false negatives.

The F1-score is the harmonic mean of precision and recall, providing a balanced metric when both false positives and false negatives are important:
\[
\text{F1-score} = \frac{2 \cdot \text{Precision} \cdot \text{Recall}}{\text{Precision} + \text{Recall}}
\]

\textbf{Overall accuracy} is computed across all three classes combined as:
\[
\text{Accuracy} = \frac{\sum_{i=1}^{3} TP_i}{N}
\]
where \(TP_i\) is the number of correctly classified samples for class \(i\), and \(N\) is the total number of samples across all classes.  
Since our benchmark includes \(66\) examples per class from the test set, \(N = 66 \times 3 = 198\).

\subsection{Zero-shot Performance}

As anticipated, zero-shot performance for \texttt{CR1} was very poor, with 0 precision and recall for Collision and Near-Collision detection with reasoning turned off. With reasoning enabled it still had 0 precision and recall for Collision cases but this time could detect some Near-Collision cases with poor precision and recall as depicted in \textbf{Table~\ref{tab:final_model_comparison}}. The majority of predictions defaulted to ``Normal Driving'' regardless. NVILA and Qwen2.5-VL-7B showed similar results with extremely poor Collision and Near-Collision metrics. These poor results show a huge domain gap that can be reduced by post-training these models using domain-specific datasets.

\subsection{Fine-tuning Performance Overview}

We evaluate \texttt{CR1} and \texttt{NVILA} in both zero-shot and fine-tuned settings. As shown in Table~\ref{tab:final_model_comparison}, all models exhibit poor zero-shot performance, particularly on the minority classes (Collision, Near-Collision), due to strong bias toward ``Normal Driving'' predictions.

After applying our proposed fine-tuning pipeline, which includes multimodal supervision (MCQs, captions, VQA, and reasoning traces) and class-balanced sampling, all models achieve significant improvements:

CR1 achieves 0.947 precision / 0.545 recall for Collision detection and an overall accuracy of 77.27\%, corresponding to an $\approx$ +42 improvement over its zero-shot baseline.
NVILA-8B, despite lacking a reasoning head, reaches 0.901 precision / 0.696 recall for Collisions and an overall accuracy of 86.4 \%, confirming its strong visual representation quality when paired with our multimodal dataset.

We provide detailed ablation analyses in the following sections to isolate the impact of each design choice.

\subsection{Ablation Studies and Key Findings}

Our ablation studies confirmed the importance of several factors related to the frame rate, video resolution, learning rate, video duration, data mixtures.

\begin{table}[h!]
  \centering
  \caption{Effect of Frame Rate (FPS)}
  \small
  \begin{tabular}{l|ccc|c}
    \toprule
    \makecell{\textbf{FPS}\\\textbf{}} & \multicolumn{3}{c|}{\textbf{F1-scores}} & \makecell{\textbf{Overall}\\\textbf{Accuracy} $\uparrow$} \\
    & \textbf{C} $\uparrow$ & \textbf{NC} $\uparrow$ & \textbf{N} $\uparrow$ & \\
    \midrule
    1 & 0.341 & 0.029 & 0.536 & 40.91 \\
    15 & 0.537 & 0.634 & 0.733 & 65.66 \\
    30 & \textbf{0.640} & \textbf{0.757} & \textbf{0.849} & \textbf{76.26} \\
    \bottomrule
  \end{tabular}
  \label{tab:fps_comparison}
\end{table}

\subsubsection{Effect of FPS}

We evaluate the effect of temporal resolution by training the model with different frame rates. As shown in Table~\ref{tab:fps_comparison}, higher FPS yields substantial gains in per-class F1-scores and overall accuracy.

At 1 FPS, the model performs poorly on temporally brief anomalies, especially Near-Collisions, with an overall accuracy of only 40.91\%. Increasing to 15 FPS significantly improves Near-Collision detection and raises accuracy to 65.66\%. At 30 FPS, the model achieves its best performance, with an overall accuracy of 76.26\% and strong F1-scores across all classes. These results confirm that high temporal fidelity is critical for detecting short, fine-grained anomalies in real-world driving videos.

\begin{table}[h!]
  \centering
  \caption{Effect of Model Size.}
  \small
  \begin{tabular}{l|c}
    \toprule
    \textbf{Model Size} & \makecell{\textbf{Overall}\\\textbf{Accuracy} $\uparrow$} \\
    \midrule
    NVILA-8B & \textbf{83.83} \\
    NVILA-33B & 80.30 \\
    \bottomrule
  \end{tabular}
  \label{tab:nvila_model_size_comparison}
\end{table}

\subsubsection{Effect of Model Size}

Our experiments reveal that varying the model size between NVILA-8B and NVILA-33B results in negligible differences in overall accuracy (Table~\ref{tab:nvila_model_size_comparison}). In fact, the 8B model slightly outperforms the 33B model (83.83\% vs.\ 80.30\%), suggesting that, for the current dataset scale, larger model capacity does not necessarily translate into performance gains. This indicates that the dataset size may be the limiting factor, as larger models typically require substantially more training data to realize their full potential. Consequently, while both models achieve comparable results, scaling the dataset could enable the 33B model to surpass the smaller variant in future experiments.

\begin{table}[h!]
  \centering
  \caption{Effect of learning rate.}
  \small
  \begin{tabular}{l|ccc|c}
    \toprule
    \makecell{\textbf{Learning}\\\textbf{Rate}} & \multicolumn{3}{c|}{\textbf{F1-scores}} & \makecell{\textbf{Overall}\\\textbf{Accuracy} $\uparrow$} \\
    & \textbf{C} $\uparrow$ & \textbf{NC} $\uparrow$ & \textbf{N} $\uparrow$ & \\
    \midrule
    1e-06 & 0.567 & 0.271 & 0.674 & 55.55 \\
    2e-05 & \textbf{0.646} & 0.725 & 0.802 & 73.73 \\
    1e-05 & 0.640 & \textbf{0.757} & \textbf{0.849} & \textbf{76.26} \\
    \bottomrule
  \end{tabular}
  \label{tab:learning_rate_comparison}
\end{table}

\subsubsection{Effect of learning rate}

We ablate different learning rates to identify the optimal setting for SFT. As shown in Table~\ref{tab:learning_rate_comparison}, both $1 \times 10^{-5}$ and $2 \times 10^{-5}$ outperform the lower $1 \times 10^{-6}$ baseline by a large margin. The best overall accuracy (76.26\%) and strongest F1-scores across most classes are achieved with $1 \times 10^{-5}$. This underscores the importance of learning rate tuning, especially for long-tail event classes.

\begin{table}[h!]
  \centering
  \caption{Impact of Video Duration and Resolution.}
  \small
  \begin{tabular}{l|l|ccc|c}
    \toprule
    \makecell{\textbf{Video}\\\textbf{Duration}} & \makecell{\textbf{Spatial}\\\textbf{Resolution}} & \multicolumn{3}{c|}{\textbf{F1-score} $\uparrow$} & \makecell{\textbf{Overall}\\\textbf{Accuracy} $\uparrow$} \\
    & & \textbf{C} & \textbf{NC} & \textbf{N} & \\
    \midrule
    6 sec & 32x4 & 0.500 & 0.702 & 0.820 & 70.20 \\
    6 sec & 128x32 & 0.640 & \textbf{0.757} & 0.849 & 76.26 \\
    4 sec & 128x32 & 0.672 & 0.744 & 0.845 & 76.26 \\
    4 sec & 192x48 & \textbf{0.711} & 0.748 & \textbf{0.857} & \textbf{77.77} \\
    \bottomrule
  \end{tabular}
  \label{tab:video_duration_spatial_res_comparison}
\end{table}

\subsubsection{Effect of Video Duration and Spatial Resolution}

We jointly ablate spatial resolution and video duration to evaluate their effect on driving event classification. As shown in Table~\ref{tab:video_duration_spatial_res_comparison}, both increasing resolution and shortening duration lead to performance gains, up to practical limits imposed by GPU memory and model context size.

Moving from 32$\times$4 to 128$\times$32 resolution at 6 seconds improves accuracy from 70.20\% to 76.26\%. Further gains are achieved by shortening the input duration to 4 seconds while increasing resolution to 192$\times$48, resulting in the best overall accuracy of 77.77\%.

These results suggest that higher spatial resolution provides critical fine-grained visual cues, and shorter durations help reduce context dilution, allowing the model to focus more effectively on anomalous frames at higher resolution.

\subsubsection{Effect of Data Mixtures}

\label{sec:data-mixing}

A crucial aspect of our post-training framework is the strategic use and combination of various supervision signals. Our ablation studies reveal that the composition, scale, and balance of the training data mixture significantly impact model performance.

\begin{table}[h!]
  \centering
  \caption{Effect of Data Balancing.}
  \small
  \begin{tabular}{l|ccc|c}
    \toprule
    \makecell{\textbf{Data}\\\textbf{Distribution}} & \multicolumn{3}{c|}{\textbf{F1-scores}} & \makecell{\textbf{Overall}\\\textbf{Accuracy} $\uparrow$} \\
    & \textbf{C} $\uparrow$ & \textbf{NC} $\uparrow$ & \textbf{N} $\uparrow$ & \\
    \midrule
    Unbalanced & 0.00 & 0.603 & 0.748 & 56.57 \\
    Balanced & \textbf{0.640} & \textbf{0.757} & \textbf{0.849} & \textbf{76.26} \\
    \bottomrule
  \end{tabular}
  \label{tab:f1_comparison}
\end{table}

\paragraph{Impact of Data Balancing}

The raw Nexar dataset is heavily imbalanced. Fine-tuning on this distribution yields low F1-scores for critical anomaly classes especially Collision class (F1 = 0.00) despite moderate performance on Normal Driving (Table~\ref{tab:f1_comparison}).

To mitigate this, we upsampled Collision and Near-Collision classes by 15$\times$ and 2$\times$ respectively, producing a balanced dataset of $\sim$68,000 samples. Fine-tuning on this mixture improved overall accuracy by +19.7\% and drastically increased F1 for Collision class (from 0.00 to 0.640) and Near-Collisions (from 0.603 to 0.757), confirming that balanced class distributions are essential for anomaly detection.

Additionally, our studies showed that a sufficient scale of data is also essential; even with a balanced distribution, using a very limited number of samples (e.g., 1,000 or 3,000 per class) resulted in a substantial degradation in performance compared to the full 68,000-sample balanced dataset.

\begin{table}[h!]
  \centering
  \caption{Accuracy for NVILA model with incremental data inclusion.}
  \small
  \begin{tabular}{l|ccc|c}
    \toprule
    \textbf{Model} & \textbf{MCQ} & \textbf{Caption} & \textbf{VQA} & \textbf{Accuracy} $\uparrow$ \\
    \midrule
    \multirow{3}{*}{NVILA} 
    & \cmark &        &        & 80.30 \\
    & \cmark & \cmark &        & 83.83 \\
    & \cmark & \cmark & \cmark & \textbf{86.36} \\
    \bottomrule
  \end{tabular}
  \label{tab:nvila_single_column}
\end{table}

\begin{table*}[t]
  \centering
  \caption{\textbf{Effect of Chain-of-Thought Reasoning Supervision.} Reasoning MCQs target classification alignment; Reason×2 doubles that data; Reason-VQA enables open-ended scene reasoning. Combining both yields the highest classification and reasoning-mode performance.}
  \normalsize
  \begin{tabular}{l|cccccc|cc}
    \toprule
    \textbf{Model} 
    & \textbf{MCQ} 
    & \textbf{Caption} 
    & \textbf{VQA} 
    & \makecell{\textbf{Reasoning }\\\textbf{MCQ}} 
    
    & \makecell{\textbf{Reasoning }\\\textbf{MCQ x2}} 
    & \makecell{\textbf{Reasoning }\\\textbf{VQA}} 
    & \makecell{\textbf{Accuracy}\\\textbf{(Reasoning Off \xmark)}} 
    & \makecell{\textbf{Accuracy}\\\textbf{(Reasoning On \checkmark)}} \\
    \midrule
    \multirow{4}{*}{\textbf{CR1}} 
    & \checkmark & \checkmark & \checkmark & - & - & - & 74.75 & n/a \\
    & \checkmark & \checkmark & \checkmark & \checkmark & - & - & 75.25 & 52.75 \\
    & \checkmark & \checkmark & \checkmark & \checkmark & \checkmark & - & 73.73 & 61.11 \\
    & \checkmark & \checkmark & \checkmark & \checkmark & \checkmark & \checkmark & \textbf{77.27} & \textbf{63.13} \\
    \bottomrule
  \end{tabular}
  \label{tab:reasoning_accuracy_comparison}
\end{table*}

\paragraph{Contribution of Caption Augmentations}

We evaluated the impact of incorporating various captioning data on model performance:
\begin{itemize}
    \item \textbf{Metadata-Derived Captions:} Using the structured captions generated from Nexar metadata proved effective in providing the model with concrete, factual information about the driving scenario. This helped ground the VLM's understanding of key details like weather, impact side, and vehicle roles, leading to improved classification accuracy, particularly for Near-Collision and Collision events.
    \item \textbf{VLM-Generated Captions:} The addition of Gemini-generated captions, which offer more free-form, descriptive summaries, further enhanced the model's performance. By exposing the model to more natural language descriptions of the same event, we observed improvements in its ability to generalize to new, unseen scenarios. Our experiments showed that a subset of approximately 10,000 Gemini captions yielded the best results when combined with other data types, suggesting a sweet spot for data diversity without introducing noise.
\end{itemize}

As shown in Table~\ref{tab:nvila_single_column}, adding captions increased NVILA’s accuracy from 80.30\% (MCQ only) to 83.83\%.

\paragraph{Impact of Question-Answering Data.}

VQA supervision, generated from detailed metadata captions, enhanced the model’s ability to reason about situational and causal elements. NVILA’s accuracy increased from 83.83\% to 86.36\% when VQA was added (Table~\ref{tab:nvila_single_column}). These structured Q\&A examples prompt the model to infer blame, event type, and intent, contributing to more accurate and interpretable classifications.

\paragraph{Role of Reasoning and Think Trace Data}

We design two distinct forms of reasoning supervision to preserve and enhance the model’s chain-of-thought (CoT) capabilities during fine-tuning: (i) \textbf{Reasoning MCQs} that target our classification task, and (ii) \textbf{Reasoning-VQA} that elicit general scene understanding through open-ended question-answering.

\begin{itemize}
    \item \textbf{Reasoning MCQs (Classification-Targeted):}  
    For each driving clip, we prompt DeepSeek-R1 with our standard 3-way MCQ along with structured and VLM-generated captions to generate a step-by-step reasoning trace and final answer. If the predicted label matches the ground truth, the explanation between \texttt{<think>} tokens is retained as a valid reasoning trace for CoT SFT.
 
    To evaluate the impact of reasoning scale and due to limited reasoning data, we double the amount of Reasoning MCQ data during SFT.

    \item \textbf{Reasoning VQA (Open-Ended Reasoning):}  
    We also generate challenging, open-ended VQA prompts that require causal, temporal, or spatial reasoning. For each example, we ask DeepSeek-R1 to generate a question, correct answer, and a think trace justification triplet, using both structured and VLM captions as input.
\end{itemize}

\paragraph{Results.}  
As shown in Table~\ref{tab:reasoning_accuracy_comparison}, adding Reasoning MCQs to the base model improves classification accuracy slightly (74.75\% → 75.25\%) but, more importantly, restores the model’s ability to generate coherent reasoning traces.

We evaluate \textbf{reasoning-mode accuracy} by prompting the model to answer with \emph{step-by-step reasoning followed by a final answer}. In contrast, the standard classification setting prompts for a direct answer without reasoning. Before adding CoT supervision, the model fails to respond with reasoning traces, even when prompted, indicating catastrophic forgetting of its pretrained CoT abilities. Adding Reasoning MCQs enables the model to recover reasoning-mode accuracy (52.75\%) and improves interpretability. Doubling the Reasoning MCQ data (×2) increases reasoning accuracy to 61.11\%. Finally, combining Reasoning MCQ with open-ended Reasoning-VQA supervision yields the best results:  
\textbf{77.27\% classification accuracy} and \textbf{63.13\% accuracy} (with reasoning enabled during inference).

These findings show that explicit CoT supervision not only improves classification performance but also enables robust and interpretable reasoning in downstream inference.

While reasoning-mode accuracy improves significantly with CoT supervision, it does not yet surpass classification accuracy. This is likely due to limited scale and diversity in the reasoning supervision, which may cause the model to overfit to the classification objective. We leave the development of larger and more robust reasoning datasets, along with refined CoT prompting pipelines, as future work to further improve reasoning-mode generalization.

\subsubsection{Final Results}

Our best-performing model for CR1 is fine-tuned with a combination of MCQs, structured and LLM-generated captions, VQA pairs, and two forms of chain-of-thought supervision: classification-aligned Reasoning MCQs and open-ended Reasoning VQA.

This model achieves an overall classification accuracy of \textbf{77.27\%} and a reasoning-mode accuracy of \textbf{63.13\%} when prompted to generate think traces during inference (Table~\ref{tab:reasoning_accuracy_comparison}). It also yields strong per-class F1-scores of 0.69 (Collision), 0.758 (Near-Collision), and 0.843 (Normal Driving), demonstrating robust performance across the full spectrum of driving events.

These results represent a significant improvement over models trained with standard classification data alone, highlighting the importance of multimodal and reasoning-aligned supervision in post-training large VLMs for safety-critical tasks.

For comparison, NVILA-8B, a non-reasoning model, achieved the highest overall classification accuracy (86.4 \%) when fine-tuned on the same multimodal dataset without reasoning supervision.
Since NVILA lacks a chain-of-thought head, we restricted its training to MCQ, caption, and VQA supervision only.
This comparison underscores the complementarity between reasoning-rich models (CR1) and efficient non-reasoning architectures (NVILA), the former excelling in interpretability and causal reasoning, the latter in pure classification efficiency.

\section{Discussion}

The substantial improvement in performance from near-zero for zero-shot to high classification and reasoning accuracy demonstrates the value of post-training VLMs for real-world tasks like anomalous driving behavior detection. This work highlights a promising direction for adapting general-purpose VLMs to safety-critical domains. This framework bridges vision-language understanding and embodied reasoning, aligning with the emerging paradigm of physical AI and world foundation models for real-world autonomy.

\subsection{Limitations and Future Work}
Despite these gains, several limitations remain, particularly in terms of generalization and reasoning fidelity. We outline key areas for improvement below.

\subsubsection{Instruction Following and Scene Understanding}

Training exclusively on MCQ classification tasks was found to reduce the model’s ability to follow open-ended instructions, such as captioning or question answering. To preserve general-purpose capabilities while maintaining strong classification performance, we propose the following augmentations:

\begin{itemize}
    \item \textbf{Short QA Generation (LLaMA-3.1-70B):} 
    
    To enhance instruction following and generate open-ended short QA, detailed, metadata-derived captions can serve as input to LLaMA-3.1-70B ~\cite{grattafiori2024llama3herdmodels} for generating short question-answer pairs, capturing essential scene and incident details, such as vehicle role, motion, impact side, and weather. These can help preserve generic scene understanding of the model to understand important questions about the driving scenarios for better generalization.

    \item \textbf{Augmented Caption Generation:}  
    In addition to structured captions, we can use LLaMA-3.1-70B to generate varied, descriptive augmentations that introduce linguistic diversity. This enables richer natural language exposure during fine-tuning and supports better downstream generalization.
\end{itemize}

\subsubsection{Causal Reasoning and Think Trace Generation}

While adding CoT reasoning data substantially improved reasoning-mode performance, a gap still remains between classification and reasoning accuracy. We attribute this to both scale limitations and data quality bottlenecks. To further improve CoT performance, we propose:

\begin{itemize}
    \item \textbf{Improved Think Trace Quality:}  
    Currently, only $\sim$33\% of reasoning VQA samples result in valid think traces where the LLM’s predicted answer matches the ground truth. For the remaining samples, we discard the trace. We plan to address this by training/prompt-engineering a secondary LLM to revise and align incorrect traces using structured captions, ground truth labels, and generated reasoning as inputs.

    \item \textbf{External Reasoning Datasets:}  
    To improve generic physical reasoning beyond our task-specific data, we will integrate broader sources such as the understanding and reasoning subsets of the Physical Common Sense VQA dataset used in CR1 fine-tuning.

    \item \textbf{Reinforcement Learning after CoT Fine-Tuning:}  
    Inspired by the RL stage in CR1 and recent Long-RL strategies, we plan to explore reinforcement learning after CoT supervised fine-tuning. This could help further stabilize reasoning trace generation, align reasoning quality with downstream performance, and mitigate overfitting to static supervision data.
    
\end{itemize}

\subsubsection{Expansion and Generalization}

In future work, we plan to:
\begin{itemize}
    \item Explore a broader range of driving behaviors (e.g., red-light violations, stop-sign infractions).
    \item Experiment with additional model architectures and hyperparameter tuning.
    \item Evaluate in simulated or closed-loop driving environments for policy transfer.
\end{itemize}

These improvements aim to further enhance the reasoning ability, instruction-following robustness, and downstream safety of post-trained VLMs for autonomous systems.

\subsubsection{Ethical Considerations}

VLM-AutoDrive is designed to enhance interpretability and safety analysis, not to replace human judgment. We acknowledge potential dataset biases and privacy risks inherent in real-world dashcam data and emphasize transparent, auditable reasoning outputs to promote responsible deployment in automotive applications.

\section{Conclusion}

This paper presents a modular post-training framework for adapting general-purpose VLMs to detect safety-critical driving events from real-world dashcam footage. Zero-shot evaluations showed that pretrained models like NVIDIA’s CR1 perform poorly on long-tail events, heavily favoring “Normal Driving” predictions.

Our solution integrates a scalable data pipeline that synthesizes diverse supervision signals, including metadata-derived captions, LLM-generated visual descriptions, VQA pairs, and chain-of-thought (CoT) reasoning prompts with think traces. These annotations are used for supervised fine-tuning.

Experiments on Nexar dashcam videos demonstrate substantial gains in classification and reasoning accuracy. For example, fine-tuning CR1 improved Collision detection precision from 0.0 to 0.947 and recall from 0.0 to 0.545. Near-Collision detection also improved significantly, with precision and recall reaching 0.696 and 0.833, respectively. Binary anomaly detection accuracy reached \textbf{87.9\%}.

Our ablations underscore the importance of high frame rates (30 FPS), fine spatial resolution, and a balanced data mixture that includes both structured and reasoning-rich supervision. The modularity of the VLM-AutoDrive pipeline also enables extensibility to new event types, making it a practical foundation for deploying VLMs in safety-critical automotive systems.

We hope VLM-AutoDrive serves as a foundation for reasoning-centric perception systems that can extend to broader physical reasoning domains beyond driving.

\section*{Acknowledgments}
We acknowledge Nexar for providing access to commercially licensed driving data used in this study. The data was obtained through a paid license. All methodology, analysis, and conclusions in this work were developed internally by NVIDIA. We also thank Will Kim and Ekram Mukbil for assistance with data access and internal coordination.


{\small
\bibliographystyle{abbrv} 
\bibliography{oldbib}
}

\clearpage

\clearpage
\onecolumn
\appendix
\section{Appendix}

\renewcommand{\thefigure}{A.\arabic{figure}}
\setcounter{figure}{0}

\subsection{Understanding VQA Generation.}
\begin{center}
    \includegraphics[width=\textwidth]{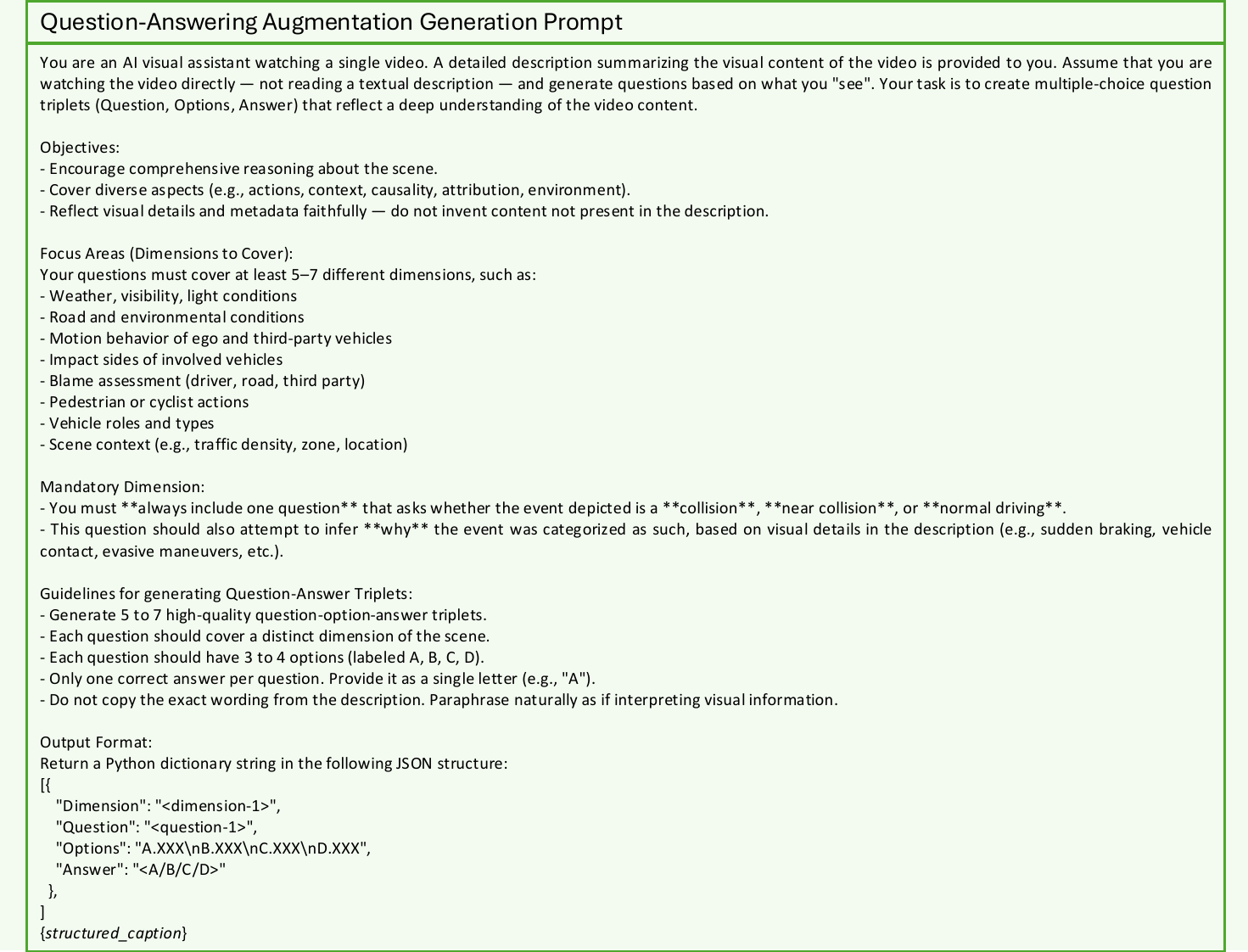}
    \captionof{figure}{Prompt used to generate understanding VQA from LLaMA3.1-70B for enhancing generic event understanding capability, given programmatically generated structured-captions as input.}
    \label{fig:understanding_VQA}
\end{center}

\clearpage

\subsection{Reasoning VQA Generation.}
\begin{center}
    \includegraphics[width=\textwidth]{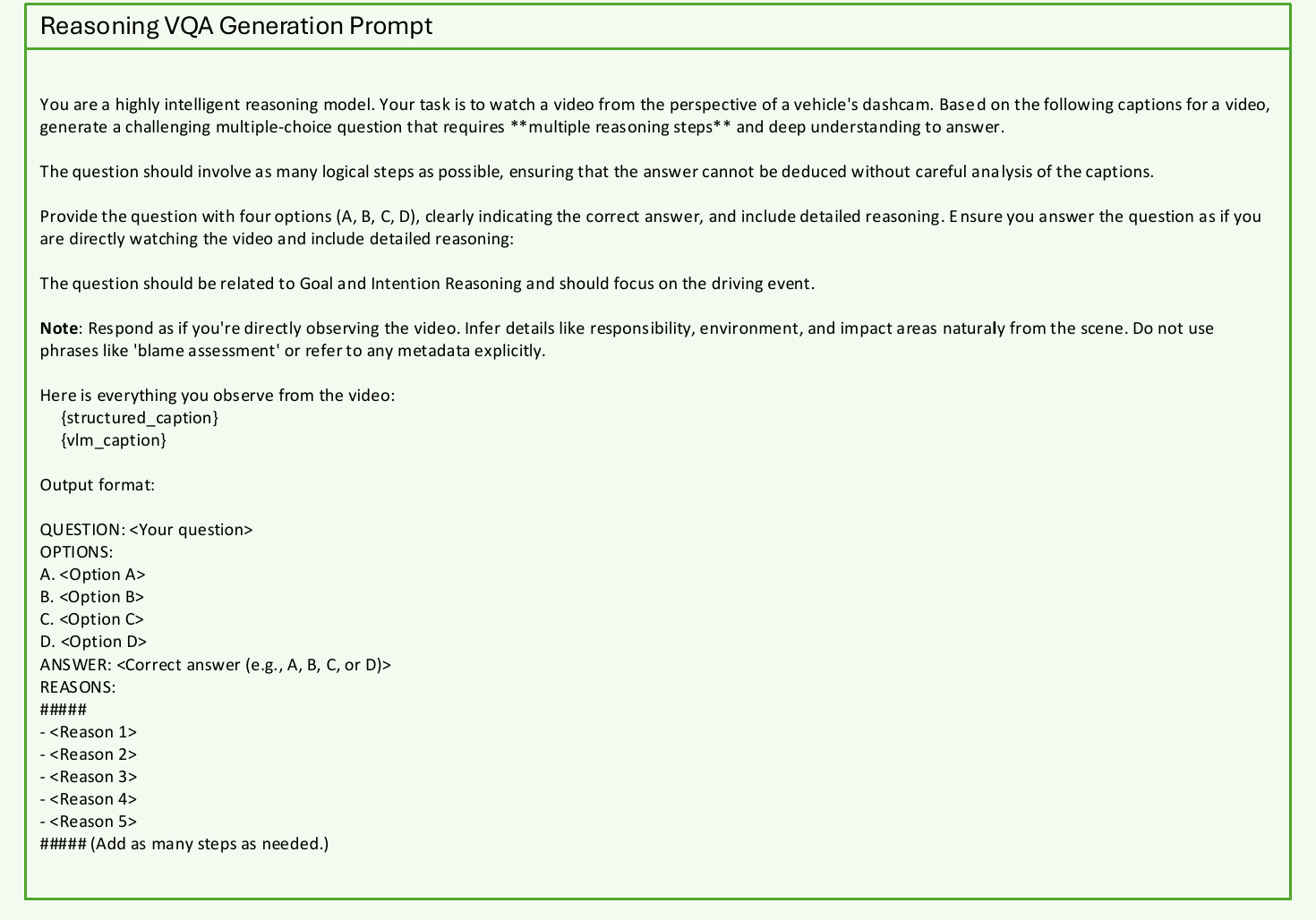}
    \captionof{figure}{Prompt used to generate reasoning VQA from DeepSeek-R1 for enhancing generic reasoning capability, given programmatically generated structured-captions and VLM generated scene description as inputs.}
    \label{fig:think_trace_general_VQA}
\end{center}

\clearpage

\subsection{Thinking Trace Generation.}
\begin{center}
    \includegraphics[width=\textwidth]{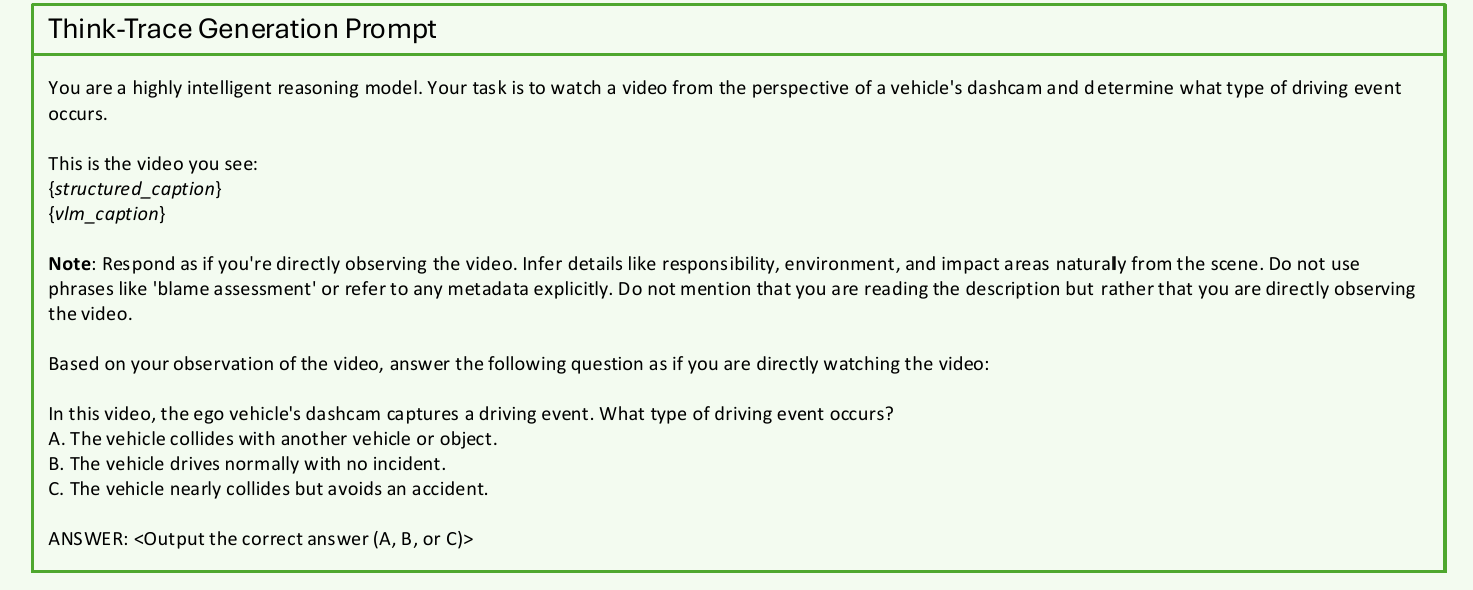}
    \captionof{figure}{Prompt used to generate thinking traces from DeepSeek-R1, given programmatically generated structured-captions, VLM generated scene description and MCQ as inputs.}
    \label{fig:chunking-full-1}
\end{center}

\end{document}